\documentclass{article}

 \usepackage[nonatbib]{icbinb_neurips_2020}

\usepackage[utf8]{inputenc} % allow utf-8 input
\usepackage[T1]{fontenc}    % use 8-bit T1 fonts
\usepackage{hyperref}       % hyperlinks
\usepackage{url}            % simple URL typesetting
\usepackage{booktabs}       % professional-quality tables
\usepackage{amsfonts}       % blackboard math symbols
\usepackage{nicefrac}       % compact symbols for 1/2, etc.
\usepackage{microtype}      % microtypography
\usepackage{amsmath}
\usepackage{xcolor}
\usepackage{graphicx}
\usepackage{subfigure}
\graphicspath{{figures/}}

\title{A Worrying Analysis of Probabilistic Time-series Models for Sales Forecasting}

 \author{%
   Seungjae Jung\thanks{All authors contributed equally to this research. The authors are sorted by alphabetical order} \\
   NAVER CLOVA\\
   \texttt{seung.jae.jung@navercorp.com} \\
   \And
   Kyung-Min Kim\footnotemark[1]\\
   NAVER CLOVA, NAVER AI LAB \\
   \texttt{kyungmin.kim.ml@navercorp.com} \\
   \AND
   Hanock Kwak\footnotemark[1]\\
   NAVER CLOVA\\
   \texttt{hanock.kwak2@navercorp.com} \\
   \And
   Young-Jin Park\footnotemark[1]\\
   NAVER CLOVA, NAVER AI LAB \\
   \texttt{young.j.park@navercorp.com} \\
 }

\begin{document}

\maketitle

\begin{abstract}
  Probabilistic time-series models become popular in the forecasting field as they help to make optimal decisions under uncertainty. 
  Despite the growing interest, a lack of thorough analysis hinders choosing what is worth applying for the desired task.
  In this paper, we analyze the performance of three prominent probabilistic time-series models for sales forecasting.
  To remove the role of random chance in architecture's performance, we make two experimental principles;
  1) Large-scale dataset with various cross-validation sets.
  2) A standardized training and hyperparameter selection.
  %We construct 
  %a real-world benchmark dataset containing 6,032 time-series of 1,725 time-steps.
  %Our dataset has more diverse levels of difficulty than publicly available datasets to make learning architecture unbiased toward a specific time-series property.
  %We compare the performances of models using ten cross-validation splits.
  %All comparative models including four baselines search hyperparameter space using 80 random trials or 36 heuristic strategies guided by an ablation study.
  The experimental results show that a simple Multi-layer Perceptron and Linear Regression outperform the probabilistic models on RMSE without any feature engineering.
  Overall, the probabilistic models fail to achieve better performance on point estimation, such as RMSE and MAPE, than comparably simple baselines.
  We analyze and discuss the performances of probabilistic time-series models.
\end{abstract}

\section{Introduction}
Sales forecasting is an essential task in logistic companies as well as retail business concerning resource optimization.
Probabilistic time-series models are becoming increasingly important in sales forecasting as it helps automate optimal decision making under uncertainty.
There have been many studies on the probabilistic time-series models \cite{adversarial, copula}, and big IT companies have already commercialized basic models through cloud services \cite{aws, oracle}.
They are being applied to various kinds of sequential data such as images \cite{attentive}, traffic, and finance \cite{salinas2020deepar, price}.

Despite their tremendous studies and usages, it is hard to precisely evaluate the progress being made due to the lack of rigorous empirical evaluation procedures.
Researchers often evaluate their models with different training procedures;
such factors, including dataset split, data normalization, model hyperparameters, evaluation metrics, and a choice of baselines, can affect the performance of learning architecture, which may unfairly benefit the new model \cite{lipton2018unfair}.
As such, we also empirically experienced that prominent probabilistic time-series models do not work effectively especially for other datasets not used in the original papers.

In this paper, we carefully study those concerns with the following main research question: To what extent the publicly used probabilistic time-series models are better than simple baselines in forecasting performance?
To answer the question, we choose three probabilistic time-series models such as DeepAR, DeepState, and Prophet, after scanning the widely used cloud services and referenced reproducible papers \cite{salinas2020deepar, rangapuram2018deep, taylor2018forecasting}. 
Moreover, the paper aims to thoroughly discuss the potential factors that can undermine the performance of probabilistic models in other datasets as well.

To eliminate the possibility of changes in performance due to external factors, we make following experimental principles.
1) Large-scale dataset with various cross-validation splits. We construct a real-world sales dataset named EC dataset consisting of 6,032 time-series of 1,725 time steps, each representing daily sales for a shop on an e-commerce website. We use ten cross-validation splits. Spectral entropy density analysis \cite{entropy} shows that the entropy density distribution of EC dataset is much more balanced than that of publicly available datasets, preventing learning architecture from biasing toward a specific time-series property.
2) A standardized training and hyperparameter selection procedure for all models including four baselines. In such setting, the differences in performance can be attributed to architectures, not other factors. 
For each split, all comparative models search hyperparameter space using 72 randomized trials or heuristic trials guided by an ablation study.
Section \hyperref[sec:setup]{3.3} describes the experimental settings in more detail.
% 3) Measurement of only the time series processing ability.
% \textcolor{red}{
% We exclude explanatory features, e.g., day-of-week, in the prediction structures of all models.
% }
%This setup allows us to assess the forecasting performance of different models more accurately and does not just select the model that overfits one fixed test set.

The experimental results show that a simple Multi-layer Perceptron (MLP) comfortably beats all competitive probabilistic models on RMSE and MAPE metrics.
A Linear Regression follows the MLP on RMSE metric.
Overall, the probabilistic models fail to achieve better performance on point estimation than simple non-probabilistic baselines.
We include further analysis and discussion in Section \hyperref[sec:analysis]{4} and \hyperref[sec:discussion]{5}. We would like to make a disclaimer that point estimation such as RMSE and MAPE are not the only way to judge the performance of probabilistic time-series models.
Learning and representing a probability distribution of output is an important research direction of the field.
However, rigorous empirical evaluation of the models is critical to understanding the model's strengths and weaknesses to use them in both commerce and academia.

\section{Time-series Models}

% This section describes the general probabilistic forecasting problem for time-series data.
% Formally speaking, 
Let a set of $N$ univariate time series, $Z = \{ z_{i, 1:T} \}_{i=1}^{N}$ where $z_{i, 1:T} = [z_{i, 1}, z_{i, 2}, \cdots, z_{i, T}]$ and $z_{i, t} \in \mathbb{R}$ represents the value of the $i$-th time series at time $t$ \footnote{The time series of variable length can be applied without loss of generality, but this paper fixed the length for the sake of simplicity of the implementation.}.
Additionally, consider a set of corresponding covariates (i.e., feature vectors), $X =\{ \mathbf{x}_{i, 1:T} \}_{i=1}^{N}$ where $\mathbf{x}_{i, 1:T} = [\mathbf{x}_{i, 1}, \mathbf{x}_{i, 2}, \cdots, \mathbf{x}_{i, T}]$ and $\mathbf{x}_{i, t} \in \mathbb{R}^d$.
The goal of probabilistic sales forecasting problem is to predict the conditional probability distribution of future time series, $z_{i, T+1:T+\tau}$, given the past observation and covariates:
\begin{equation}
    p(z_{i, T+1:T+\tau} | z_{i, 1:T}, \mathbf{x}_{i, 1:T+\tau}) .
\end{equation}

Following subsections briefly describe prominent sales forecasting models and how they predict the future time series.
Please refer to supplementary materials for more details.
% In this paper, we focus on the problem regarding each time-series as independent, and omit the superscript $(i)$ for the notational simplicity for the following sections.

\subsection{DeepAR}
DeepAR \cite{salinas2020deepar} is one of the most prominent forecasting model that is based on the auto-regressive recurrent network architecture \cite{graves2013generating, sutskever2014sequence}.
DeepAR approximates the conditional distribution using a neural network as follows:
\begin{align}
    p(z_{i, T+1:T+\tau} | z_{i, 1:T}, \mathbf{x}_{i, 1:T+\tau})
    & = \prod_{t=T+1}^{T+\tau} p(z_{i, t} | z_{i, 1:t-1}, \mathbf{x}_{i, 1:T+\tau})
    = \prod_{t=T+1}^{T+\tau}  l_\theta (z_{i, t}; \mathbf{h}_{i, t}) \\
    \mathbf{h}_{i, t+1} &= h(\mathbf{h}_{i, t}, z_{i, t}, \mathbf{x}_{i, t}) \nonumber 
\end{align}
where $l_\theta$ is a likelihood function parameterized by $\theta$ and $\mathbf{h}_{i, t}$ is an output of the recurrent network, $h$.
% Note that the recurrent network $h$ is globally utilized among different time-series collections, thus the learned parameters are shared.
% To be specific, we can parameterize the model for a Gaussian likelihood distribution by using two affine functions:
% \begin{gather}
%     l_\theta(z | \mathbf{h}) = \mathcal{N}(z ; \mu_\theta(\mathbf{h}), \sigma_\theta(\mathbf{h})) \\
%     \mu_\theta(\cdot) = \mathbf{w}_{\mu}^T \mathbf{h} + b_{\mu} ,~~~~ \sigma_\theta(\cdot) = \mbox{softplus} (\mathbf{w}_{\sigma}^T \mathbf{h} + b_{\sigma}) \nonumber 
% \end{gather}
% where $\mathcal{N}(\cdot)$ is a likelihood function of normal distribution. 

\subsection{DeepState}
Deep State Space Models (DeepState) \cite{rangapuram2018deep} is a probabilistic forecasting model that fuses state space models and deep neural network.
% State space representation provides a general framework for describing the temporal structure in time-series data \cite{hamilton1994state}.
State space model introduces a latent Markovian state $\mathbf{l}_t$ encoding temporal features, and models the time-series observation via transition $p(\mathbf{l}_t | \mathbf{l}_{t-1})$ and observation model $p(z_t | \mathbf{l}_{t})$.
In particular, DeepState adopts linear Gaussian models:
\begin{align} \label{eq:SSM}
\mathbf{l}_t &= \mathbf{F}_t\mathbf{l}_{t-1} + \mathbf{g}_t \varepsilon_t & \varepsilon_t \sim \mathcal{N}(0, I) \\
z_t  &= \mathbf{a}_t^{T}\mathbf{l}_t + b_t + \sigma_t \nu_t & \nu_t \sim \mathcal{N}(0, 1) \nonumber 
\end{align}
where $\mathbf{l}_1 \sim \mathcal{N}(\boldsymbol{\mu}_0, \boldsymbol{\Sigma}_0)$ and $\Theta_0 = \{\boldsymbol{\mu}_0, \boldsymbol{\Sigma}_0\}$ are initial state parameters, while  $\Theta_t = \{ \mathbf{F}_t, \mathbf{g}_t, \mathbf{a}_t, b_t, \sigma_t\}$ are time-varying parameters of the model.
% In traditional state space model, model parameters are often regarded to be known, whereas recent neural network based framework often focuses on the problem where they are completely unknown.
DeepState estimates model parameters for each time series by using recurrent neural network $h$ which takes covariates as input:
\begin{align} \label{eq:deepstate}
    & \Theta_{i, t} = f_{\theta}(\mathbf{h}_{i, t}) ,~~~~
    \mathbf{h}_{i, t+1} = h(\mathbf{h}_{i, t}, \mathbf{x}_{i, t}) 
    & \forall t \in \{0, \cdots, T+\tau\}
\end{align}
where $f_{\theta}$ is a function mapping the output vector of the recurrent neural network to the parameters.
% , while DeepState applied simple affine transformation followed by suitable activation function (e.g., the softplus function for positive parameters).
Unlike the auto-regressive models, DeepState uses the observation values to compute the posterior distribution of latent state, $p(\mathbf{l}_{i, T} | z_{i, 1:T})$, by integrating Kalman filtering \cite{kalman1960new, welch1995introduction}.
DeepState estimate the joint distribution of future time series from the estimated posterior and state space models.

% Unlike the auto-regressive models such as DeepAR, DeepState uses the observation values to compute the posterior distribution of latent state at each time step.
% Especially, DeepState integrates Kalman filtering \cite{kalman1960new, welch1995introduction} into the model to estimate the analytical solution for the posterior: $p(\mathbf{l}_{i, T} | z_{i, 1:T})$.
% From the estimated posterior, DeepState predict the conditional joint distribution of future time-series as follows:
% \begin{align}
%     & p(z_{i, T+1:T+\tau} | z_{i, 1:T}, \mathbf{x}_{i, 1:T+\tau}) = \prod_{t=T+1}^{T+\tau} p(z_{i, t} | z_{i, 1:t-1}; \Theta_{0:t-1}^{(i)}) \nonumber \\
%     & = \int p(\mathbf{l}_{i, T} | z_{i, 1:T}; \Theta_{0:T}^{(i)}) \left[ \prod_{t=T+1}^{T+\tau} p(z_{i, t} | \mathbf{l}_{i, T}; \Theta_{i, t}) p(\mathbf{l}_{i, t} | \mathbf{l}_{i, t-1}; \Theta_{i, t}) \right] d\mathbf{l}_{T:T+\tau}^{(i)} .
% \end{align}
% Although the joint distribution is analytically tractable, it is evaluated by Monte Carlo approaches for the sake of implementation convenience following the reference paper. 

\subsection{Prophet}
Prophet\cite{taylor2018forecasting} is a modular regression model with interpretable parameters.
It is a sum of trend $g(t)$, seasonality $s(t)$, and holidays $h(t)$ components as follows: $z_{t} = g(t) + s(t) + h(t) + \epsilon_t$, where $\epsilon_t$ is a normally distributed error term.
Prophet incorporates trend changepoints in the trend model that the analysts can manually configure or the model automatically finds.
The seasonality model $s(t)$ is a general Fourier series with a regular period that we manually set.

% \subsection{N-BEATS}

% N-BEATS\cite{oreshkin2020nbeats} has backward and forward residual links and a very deep stack of neural blocks.
% Each block infers coefficients of own basis functions, and the output is a linear combination of them.
% The final output of the model is the sum of all partial forecasts.
% This is similar to the basic structure of Prophet and it can interpret time-series as  composition of intuitive functions by manually defining the basis functions.
% The doubly residual links make the forecast job of the downstream blocks easier and facilitates more \textcolor{red}{fluid gradient backpropagation}.
% N-BEATS achieved state-of-the-art performance in M4 competition as a pure ML model.

% The parameters of linear trend model are growth rate $k$, offset $m$, rate adjustments $\boldsymbol{\delta}$, and changepoints $\boldsymbol{a}(t)$ which are combined to be:
% \begin{equation}
%     g(t) = (k + \boldsymbol{a}(t)^T \boldsymbol{\delta})t + (m + \boldsymbol{a}(t)^T \boldsymbol{\gamma}).
% \end{equation}
% The correct adjustment $\boldsymbol{\gamma}$ is set to connect lines in changepoints.

% The seasonality model is a general Fourier series as follows:
% \begin{equation}
%     s(t) = \sum_{n=1}^{N}\left(a_n\cos{\left(\frac{2\pi n t}{P}\right)} + b_n\sin{\left(\frac{2\pi n t}{P}\right)}\right),
% \end{equation}
% where $P$ is the regular period that we manually set and $a_n, b_n$ are learnable parameters.

\section{Experiments}

\subsection{Dataset}
To create a testbed, we construct a large-scale real-world sales dataset named EC dataset from our e-commerce website.
As most shops in the e-commerce site tend to have sparse sales records, i.e., many zeros in the daily sales history, the models tend to rely on other features instead of time series information.
We seek to make our dataset as predictable as possible and let models to learn nature of time-series data.
To select shops with steady sales performance, we filter out shops containing zeros in the daily sales records from 2019-01-01 to 2020-09-20.
As a result, the EC dataset contains daily sales records from 6,032 different shops for 1,725 days from 2016-01-01 to 2020-09-20.
Figure 1 shows the comparison of widely used public datasets and the EC dataset in terms of forecastability.
Our dataset is more trickier than traffic and electricity but easier than Wikipedia.
Figure 2 describes the comparison of spectral entropy density \cite{entropy} for the EC dataset and others.
The entropy of our dataset is much more distributed than that of others, meaning that ours is much more balanced. 
The length of test and validation sets are all seven days, and remaining past days are training set.
We make ten cross validation splits with no intersection in the test sets.

% \begin{table}
%   \caption{Comparison of forecastability for widely used public datasets and ours. CoV denotes coefficient of variation. The higher the value, the more difficult to forecast.}
%   \label{dataset-comparison}
%   \centering
%   \begin{tabular}{llll}
%     \toprule
%     Dataset       &  Mean Spectral Entropy    & Mean CoV \\
%     \midrule
%     traffic      & 0.264    & 0.477        \\
%     electricity   & 0.263    & 0.741        \\
%     Wikipedia    & 0.837    & 2.887 \\
%     EC (ours)      &0.778    & 1.112        \\
%     \bottomrule
%   \end{tabular}
% \end{table}
% \begin{figure}[h]
% \label{entropy-density}
% \includegraphics[scale=0.4]{ec_histogram.png}
% \caption{Comparison of spectral entropy density for widely used public datasets and ours. Entropy of our dataset is much more distributed than that of others}
% \end{figure}

\begin{figure}
    \begin{minipage}{.45\linewidth}
        \centering
        \includegraphics[width=0.95\textwidth]{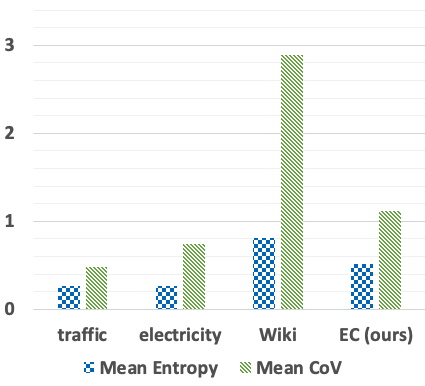}
    \end{minipage}
    \hspace{0.05\textwidth}
    \begin{minipage}{.45\linewidth}
        \centering
        \includegraphics[width=1.1\textwidth]{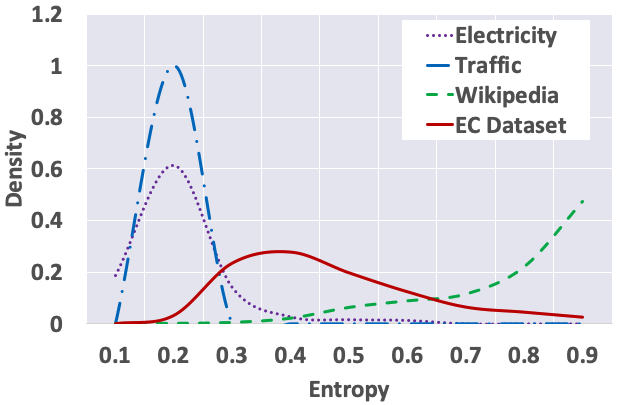} 
        %\caption{Comparison of spectral entropy density. The EC dataset allows to test varying difficulty levels.}
    \end{minipage} \\[5pt]
    \begin{minipage}[t]{.45\linewidth}
        \caption{Comparison of forecastability for EC dataset and publicly available datasets.}
    \end{minipage}
    \hfill
    \begin{minipage}[t]{.45\linewidth}
        \caption{Comparison of spectral entropy density. The EC dataset allows to test various time-series properties.}
    \end{minipage}
\end{figure}

\subsection{Baselines}
In addition, we consider four baseline models: MA (Moving Average), LR (Linear Regression), MLP (Multi-layer Perceptron) and SARIMA (Seasonal ARIMA).
MA is a simple unweighted mean of past few days.
The historical value of the time series is solely used as an input feature for all the baselines.
We try various historical length for MA, LR, and MLP.
We test all combination of $0 \leq p, d, q \leq 2$ for SARIMA models and fix seasonal length to seven.
The SARIMA show better performance than ARIMA, so we omit the results of ARIMA.

\subsection{Experimental Setups} \label{sec:setup}
For neural or hybrid architectures such as DeepAR, DeepState, LR, and MLP, we define a standard set of hyperparameters. 
% We use the same weight initialization (Xavier initialization \cite{xavier}), same optimization method (Adam with default parameters), and same learning rate strategy (random value between 0.0001 and 0.01 as initial learning rate with no learning rate decay).
We use the same weight initialization (PyTorch \cite{PyTorch2019} default initialization), same optimization method (Adam with default parameters), and search multiple learning rates (value between 0.0001 and 0.01).
We try randomly sampled 72 different hyperparameters for prophet and baselines with the same early stopping strategy (loss value as early stopping criteria and five as the number of patience).
Simiarly, we conduct the ablation studies of DeepAR and DeepState with 72 heuristically selected points.
We describe below detailed architecture configuration with tunable hyperparameter set.

\textbf{DeepAR}: 
We select GRU as RNN architecture for DeepAR since it slightly outperforms LSTM.
We test the model with 72 different hyperparameters: the number of RNN layers (3 and 5) $\times$ context length (7, 14, 28, 56, 84, and 112) $\times$ 3 different scaling strategy (mean, median or no scaling) $\times$ output distributions (negative binomial and Gaussian).
For features, we feed lagged values, normalized day-of-the-week, normalized day-of-the-month, and age (i.e., the distance to the first observation).
% We scale autoregressive inputs as discussed in DeepAR \cite{salinas2020deepar}.
% We use median scaling instead of mean scaling as median scaling shows better performance.

\textbf{DeepState}:
First of all, we adopt the same RNN architecture as in the setups of DeepAR for the sake of unity.
Same as in DeepAR experiments, we explore the model with two different RNN layers, six different context lengths, and three different scaling methods.
We further test two different state space model (SSM) architectures in Eq. \ref{eq:SSM}: fully-learnable (FL) SSM and partially-learnable (PL) SSM.
The total number of runs are 72 as in the experiments for DeepAR.
For PL SSM, we apply widely-used level-trend and day-of-the-week seasonality models, that are widely used approaches in the sales dataset as well as the default settings from GluonTS implementation \cite{alexandrov2019gluonts}.
In this case, the only learnable parameters are $\Theta_0$ and $\Theta_t = \{ \mathbf{g}_t, b_t, \sigma_t \}$.
To achieve a fair comparison, we use the same latent dimension for the FL SSM (i.e., $|\mathbf{l}_t| = 8$).
We use shop index, normalized day-of-the-week, and normalized day-of-the-month features as covariates.

\textbf{Prophet}: 
It is hard to identify all kinds of events in the EC dataset, so we omit holidays functions.
There are two kinds of trend models: linear and nonlinear saturating trend models.
We choose former one since saturating patterns are rare in the EC dataset.
In the official Prophet API, it recommends three effective hyperparameters: changepoint prior scale, seasonality prior scale, and seasonality mode.
The prior scales determines the flexibility of the trend and seasonality.
A large value is more likely to overfit on the training dataset.
The two options of seasonality mode are `additive' and `multiplicative' where the multiplicative seasonality multiplies the seasonality function by the trend model.

We report the results of best setups for each model.

\subsection{Metrics and Evaluation}
In practice, the costs are not scale-free since the prediction errors in the large shops would cause more financial damage.
We use RMSE (Root Mean Squared Error) to consider such aspects.
We also measure MAPE (Mean Absolute Percentage Error) for a scale-free metric:
\begin{equation}
    MAPE = \frac{1}{H}\sum_{i, t} \frac{|z_{i, t} - \widehat{z_{i, t}}|}{|z_{i, t}|},
\end{equation}
where $H$ is the total number of the prediction targets and $\widehat{z_{i, t}}$ is a prediction value.
In order to evaluate probabilistic prediction, we use NQL\cite{salinas2020deepar}\footnote{Note that NQL is the same metric as $\rho$-risk in \cite{salinas2020deepar}.} (Normalized Quantile Loss) that quantify the accuracy of a $\tau$-quantile $q_{i, t}^{(\tau)}$ of the predictive distribution as follows:
\begin{equation}
    NQL(\tau) = 2\frac{\sum_{i,t} [\tau \max(z_{i, t} - q_{i, t}^{(\tau)}, 0) + (1 - \tau) \max(q_{i, t}^{(\tau)} - z_{i, t}, 0)]}{\sum_{i,t} z_{i, t}}.
\end{equation}
Specifically, we compute mean of NQL(0.1) and NQL(0.9).

\section{Results and Analysis}
\label{sec:analysis}
\begin{table}
  \caption{RMSE, MAPE, and Mean NQL averaged over ten cross-validation splits. We perform experiments with different hyperparameter settings for all models and report each metrics's best mean score. Only probabilistic models have Mean NQL value. MA and LR denote Moving Average and Linear Regression respectively.}
  \label{tab:comparison}
  \centering
  \begin{tabular}{llll}
    \toprule
    Model       & RMSE      & MAPE    & Mean NQL[0.1, 0.9] \\
    \midrule
    MA          & 79.28$\pm$13.79  & 0.348$\pm$0.015  & -       \\
    MLP          & \textbf{72.23$\pm$12.60}     & \textbf{0.267$\pm$0.015}  & - \\
    LR          & 73.20$\pm$13.60  & 0.300$\pm$0.010  & -                   \\
    SARIMA      & 79.09$\pm$17.50  & 0.318$\pm$0.017  & 0.165$\pm$0.033     \\
    DeepAR      & 76.69$\pm$17.86  & 0.284$\pm$0.022  & 0.132$\pm$0.026     \\
    DeepState   & 79.10$\pm$15.83  & 0.314$\pm$0.030  & 0.152$\pm$0.020     \\
    Prophet     & 82.21$\pm$17.64  & 0.415$\pm$0.018  & 0.194$\pm$0.027     \\
    \bottomrule
  \end{tabular}
\end{table}

We report the prediction results of prominent probabilistic models and baseline models on the EC dataset in the following sections. 
The paper shows the negative results and their causes that degrade the performance of each model.
The ablation study illustrating hyperparameter's effect on each model is described in Section \ref{sec:ablation} of the Appendix with details.

\subsection{DeepAR} \label{sec:deepar_results}
As reported in Table \ref{tab:comparison}, DeepAR shows the best result on mean NQL compared to other probabilistic models.
Still, DeepAR performs worse than a simple MLP model on RMSE and MAPE metrics.
The result indicates that although DeepAR is a decent probabilistic model as widely known, it is not a remarkable point estimator, unlike our expectation.
The paper would like to address the point that autoregressive architecture such as RNN, is not the sole approach to dealing with time-series data.
As shown in the results, simple MLPs could capture richer latent time-series characteristics when the length of lagged values are appropriately adjusted.
We observe empirically that DeepAR's performance is affected unstably by which lagged value is selected; 
an autoregressive architecture may not effectively convey information from the distant past well over time, unlike ideal.

As mentioned in the reference paper \cite{salinas2020deepar}, we confirm that scaling method is a critical factor that hugely affects the results.
There is relatively little difference between the mean and median scaling, but there is a noticeable gap in performance depending on whether the scaling is applied or not, as shown in Table \ref{tab:scale}.
The scaling method not only helps the stability of the learning procedure by adjusting the range of values fed into the model, but prevents the over or under-fitting to a specific data instance by balancing the scale between time series.
The consistent results are shown in the experiments with DeepState.

\subsection{DeepState}
Primarily, DeepState fails to outperform competing models including DeepAR on any of three evaluation metrics.
The paper regard above results are mainly due to the inherent lack of representation power of DeepState.
The only factors that govern the system parameters in DeepState are the covariates as illustrated in Eq. \ref{eq:deepstate}, while the real-world tasks often do not provide a sufficient set of covariates.
For example, in sales data, ordinary information such as day-of-the-week and season is accessible, but latent features such as sales trend and promotion for each shop are generally not provided.
As such, if the system parameters are estimated only through indistinct features (i.e., without observation sequences), the inferred models are highly likely to be inaccurate.
Taken together, DeepState may function properly only when the target system is assumed to be fairly known and less volatile, like the traffic and electricity dataset reported in the reference paper \cite{rangapuram2018deep}.

We further compare the DeepState with the FL SSMs to PL SSMs to evaluate the effect of SSM architectures to the performance.
Our initial hypothesis is that FL model would give much higher performance compare to the PL due to its flexibility; remark that the only learnable parts in PL model were about the uncertainty estimators, while FL offers the flexibility in indetifying the transition and observation models.
FL mode, however, shows significantly dropped performance to the one with PL SSMs as described in Table \ref{tab:flpl}.
Moreover, the value of standard deviation calculated over hyperparmeter samples for FL model is noticeably large (i.e., the robustness of the model and the prevalent advantage of using probabilistic model are not guaranteed.).
We reason that the SSMs' extreme flexibility causes the non-identifiability problem \cite{doerr2018probabilistic, frigola2014variational} as well as the overfitting.

\subsection{Prophet}
The patterns of the time series are complicated and change dynamically over time, but Prophet follows such changes only with the trend changing.
The seasonality prior scale is not effective, while higher trend prior scale shows better performance.
There exist some seasonality patterns in the EC dataset, but these patterns are not consistent neither smooth.
Since Prophet does not directly consider the recent data points unlike other models, this can severely hurts performance when prior assumptions do not fit.
The detailed results are shown in Appendix.

\section{Discussion and Conclusion}
\label{sec:discussion}
This paper analyzes the performance and negative results of three prominent probabilistic time-series models such as DeepAR, DeepState, and Prophet, for sales forecasting by comparing them with simple baselines including Linear Regression and Multi-layer Perceptron.
For experiment, we construct a large-scale real-world benchmark dataset containing 6,032 time series of 1,725 time-steps with ten cross-validation splits.
We use the standardized training and hyperparameter selection procedure to remove the role of random chance in model's performance.

Prior to our experiment, we observe that the spectral entropy densities of widely used public datasets such as traffic, electricity, and Wikipedia are biased towards a specific point.
If a researcher tests his or her idea on only these datasets, specifically traffic or electricity, with a small number of baselines, it may unfairly benefit the new model. 
However, which has often been the case.

Another observation is that, surprisingly, a simple MLP and LR outperform all the competing probabilistic models on RMSE and MAPE metrics. 
Previous studies on probabilistic forecasting report their model performances based on the quantile loss while making a small effort to compare the point estimation performance to baselines.
Contrary to the researcher's interest, point estimation performance is essential in industries that require specific numbers, such as the number of delivery people in a logistics company. 
When this practice is combined with other aspects, i.e., not reporting on hyperparameter selection in the original paper, it is difficult to precisely assess the progress of the field.

Moreover, this paper points out that the belief that probabilistic models are considered robust is not always guaranteed.
In our experiments, for instance, DeepState with fully learnable SSMs shows high variance results, which is inconsistent with the statement from the original paper: DeepState is robust to noise.
In the case of the SSM-based model, in particular, this paper advises the future research papers to conduct research developing models that can further increase the stability of well-known probabilistic models by resolving the related issue such as non-identifiability.

Lastly, the paper suggests the future works to further consider the selection of the probability distribution of output that time-series model assumes.
Despite the wide variety of recent studies on probabilistic time-series models that combine neural networks, research about choosing proper probability distribution that fits a given dataset has not been sufficiently studied while previous works often assumed a Gaussian distribution  \cite{krishnan2015deep, karl2016deep, krishnan2016structured, ha2018adaptive, rangapuram2018deep, wang2019deep}.
By choosing the appropriate probability distribution, the bias in the objective function such as likelihood, become further reduced, and the prediction accuracy, especially the point estimation performance, can be improved.

% 종합 메세지 한 줄 딱!!!
%With a thorough evaluation procedure, prominent probabilistic time-series models fail to achieve better point estimation performances than that of comparably simple baselines.

% \begin{ack}
% Use unnumbered first level headings for the acknowledgments. All acknowledgments
% More information about this disclosure can be found at: \url{https://neurips.cc/Conferences/2020/PaperInformation/FundingDisclosure}.

% Do {\bf not} include this section in the anonymized submission, only in the final paper. You can use the \texttt{ack} environment provided in the style file to autmoatically hide this section in the anonymized submission.
% \end{ack}

\bibliographystyle{unsrt}
\bibliography{icbinb_neurips_2020}

\begin{thebibliography}{10}

\bibitem{adversarial}
Raphaël Dang-Nhu, Gagandeep Singh, Pavol Bielik, and Martin Vechev.
\newblock Adversarial attacks on probabilistic autoregressive forecasting
  models.
\newblock In {\em Proceedings of International Conference on Machine Learning
  (ICML)}, 2020.

\bibitem{copula}
David Salinasa, Michael Bohlke-Schneider, Laurent Callot, Roberto Medico, and
  Jan Gasthaus.
\newblock High-dimensional multivariate forecasting with low-rank gaussian
  copula processes.
\newblock In {\em Advances in neural information processing systems}, 2019.

\bibitem{aws}
Amazon web services (aws) - cloud computing services.
\newblock \url{https://aws.amazon.com}.

\bibitem{oracle}
Oracle retail demand forecasting cloud service.
\newblock
  \url{https://www.oracle.com/industries/retail/products/supply-chain-planning/demand-forecasting/}.

\bibitem{attentive}
H.~Kim, A.~Mnih, J.~Schwarz, M.~Garnelo, A.~Eslami, D.~Rosenbaum, O.~Vinyals,
  and Y.W. Teh.
\newblock Attentive neural processes.
\newblock In {\em Proceedings of International Conference on Learning
  Representations}, 2019.

\bibitem{salinas2020deepar}
David Salinas, Valentin Flunkert, Jan Gasthaus, and Tim Januschowski.
\newblock Deepar: Probabilistic forecasting with autoregressive recurrent
  networks.
\newblock {\em International Journal of Forecasting}, 36(3):1181--1191, 2020.

\bibitem{price}
Ye-Sheen Lim and Denise Gorse.
\newblock Deep probabilistic modelling of price movements for high-frequency
  trading.
\newblock {\em arXiv preprint arXiv:2004.01498}, 2020.

\bibitem{lipton2018unfair}
Zachary~C. Lipton and Jacob Steinhardt.
\newblock Troubling trends in machine learning scholarship.
\newblock {\em arXiv preprint arXiv:1807.03341}, 2018.

\bibitem{rangapuram2018deep}
Syama~Sundar Rangapuram, Matthias~W Seeger, Jan Gasthaus, Lorenzo Stella,
  Yuyang Wang, and Tim Januschowski.
\newblock Deep state space models for time series forecasting.
\newblock In {\em Advances in neural information processing systems}, pages
  7785--7794, 2018.

\bibitem{taylor2018forecasting}
Sean~J Taylor and Benjamin Letham.
\newblock Forecasting at scale.
\newblock {\em The American Statistician}, 72(1):37--45, 2018.

\bibitem{entropy}
Georg~M. Goerg.
\newblock Forecastable component analysis.
\newblock In {\em Proceedings of the 30th International Conference on Machine
  Learning}, 2013.

\bibitem{graves2013generating}
Alex Graves.
\newblock Generating sequences with recurrent neural networks.
\newblock {\em arXiv preprint arXiv:1308.0850}, 2013.

\bibitem{sutskever2014sequence}
Ilya Sutskever, Oriol Vinyals, and Quoc~V Le.
\newblock Sequence to sequence learning with neural networks.
\newblock In {\em Advances in neural information processing systems}, pages
  3104--3112, 2014.

\bibitem{kalman1960new}
Rudolph~Emil Kalman.
\newblock A new approach to linear filtering and prediction problems.
\newblock {\em Journal of basic Engineering}, 82(1):35--45, 1960.

\bibitem{welch1995introduction}
Greg Welch, Gary Bishop, et~al.
\newblock An introduction to the kalman filter, 1995.

\bibitem{PyTorch2019}
Adam Paszke, Sam Gross, Francisco Massa, Adam Lerer, James Bradbury, Gregory
  Chanan, Trevor Killeen, Zeming Lin, Natalia Gimelshein, Luca Antiga, Alban
  Desmaison, Andreas Kopf, Edward Yang, Zachary DeVito, Martin Raison, Alykhan
  Tejani, Sasank Chilamkurthy, Benoit Steiner, Lu~Fang, Junjie Bai, and Soumith
  Chintala.
\newblock Pytorch: An imperative style, high-performance deep learning library.
\newblock In {\em Advances in Neural Information Processing Systems 32}, pages
  8026--8037. Curran Associates, Inc., 2019.

\bibitem{alexandrov2019gluonts}
Alexander Alexandrov, Konstantinos Benidis, Michael Bohlke-Schneider, Valentin
  Flunkert, Jan Gasthaus, Tim Januschowski, Danielle~C Maddix, Syama
  Rangapuram, David Salinas, Jasper Schulz, et~al.
\newblock Gluonts: Probabilistic time series models in python.
\newblock {\em arXiv preprint arXiv:1906.05264}, 2019.

\bibitem{doerr2018probabilistic}
Andreas Doerr, Christian Daniel, Martin Schiegg, Duy Nguyen-Tuong, Stefan
  Schaal, Marc Toussaint, and Sebastian Trimpe.
\newblock Probabilistic recurrent state-space models.
\newblock {\em arXiv preprint arXiv:1801.10395}, 2018.

\bibitem{frigola2014variational}
Roger Frigola, Yutian Chen, and Carl~Edward Rasmussen.
\newblock Variational gaussian process state-space models.
\newblock In {\em Advances in neural information processing systems}, pages
  3680--3688, 2014.

\bibitem{krishnan2015deep}
Rahul~G Krishnan, Uri Shalit, and David Sontag.
\newblock Deep kalman filters.
\newblock {\em arXiv preprint arXiv:1511.05121}, 2015.

\bibitem{karl2016deep}
Maximilian Karl, Maximilian Soelch, Justin Bayer, and Patrick Van~der Smagt.
\newblock Deep variational bayes filters: Unsupervised learning of state space
  models from raw data.
\newblock {\em arXiv preprint arXiv:1605.06432}, 2016.

\bibitem{krishnan2016structured}
Rahul~G Krishnan, Uri Shalit, and David Sontag.
\newblock Structured inference networks for nonlinear state space models.
\newblock {\em arXiv preprint arXiv:1609.09869}, 2016.

\bibitem{ha2018adaptive}
Jung-Su Ha, Young-Jin Park, Hyeok-Joo Chae, Soon-Seo Park, and Han-Lim Choi.
\newblock Adaptive path-integral autoencoders: Representation learning and
  planning for dynamical systems.
\newblock In {\em Advances in Neural Information Processing Systems}, pages
  8927--8938, 2018.

\bibitem{wang2019deep}
Yuyang Wang, Alex Smola, Danielle~C Maddix, Jan Gasthaus, Dean Foster, and Tim
  Januschowski.
\newblock Deep factors for forecasting.
\newblock {\em arXiv preprint arXiv:1905.12417}, 2019.

\end{thebibliography}

\newpage
\appendix
{\huge\bf\raggedright Appendix}\\

\section{Parameterizaion of likelihood function for DeepAR}
This paper follows the parameterization methods described in the references \cite{salinas2020deepar, alexandrov2019gluonts} for $l_\theta$ of DeepAR.
To be specific, we parameterize the model for a Gaussian likelihood distribution by using two affine functions:
\begin{gather}
    l_\theta(z | \mathbf{h}) = \mathcal{N}(z ; \mu_\theta(\mathbf{h}), \sigma_\theta(\mathbf{h})) \\
    \mu_\theta(\cdot) = \mathbf{w}_{\mu}^T \mathbf{h} + b_{\mu} ,~~~~ \sigma_\theta(\cdot) = \mbox{softplus} (\mathbf{w}_{\sigma}^T \mathbf{h} + b_{\sigma}) \nonumber 
\end{gather}
where $\mathcal{N}(\cdot)$ is a likelihood function of normal distribution. 
The negative binomial distribution is parameterized as follows:
\begin{gather}
    l_\theta(z | \mathbf{h}) = \mathcal{BN}(z ; \mu_\theta(\mathbf{h}), \alpha_\theta(\mathbf{h})) = \frac{\Gamma(z + 1/\alpha)}{\Gamma(z+1)\Gamma(1/\alpha)} \Big( \frac{1}{1+\alpha\mu} \Big)^{\frac{1}{\alpha}} \Big( \frac{\alpha\mu}{1 + \alpha\mu} \Big)^z\\
    \mu_\theta(\cdot) = \mbox{softplus} ( \mathbf{w}_{\mu}^T \mathbf{h} + b_{\mu}) ,~~~~ \alpha_\theta(\cdot) = \mbox{softplus} (\mathbf{w}_{\alpha}^T \mathbf{h} + b_{\alpha}) \nonumber 
\end{gather}

\section{Estimation of joint distribution of future time series for DeepState}
Unlike the auto-regressive models such as DeepAR, DeepState uses the observation values to compute the posterior distribution of latent state at each time step.
Especially, DeepState integrates Kalman filtering \cite{kalman1960new, welch1995introduction} into the model to estimate the analytical solution for the posterior: $p(\mathbf{l}_{T}^{(i)} | z_{1:T}^{(i)})$.
From the estimated posterior, DeepState predict the conditional joint distribution of future time series as follows:
\begin{align}
    & P(z_{T+1:T+\tau}^{(i)} | z_{1:T}^{(i)}, \mathbf{x}_{1:T+\tau}^{(i)}) = \prod_{t=T+1}^{T+\tau} p(z_t^{(i)} | z_{1:t-1}^{(i)}; \Theta_{0:t-1}^{(i)}) \nonumber \\
    & = \int p(\mathbf{l}_{T}^{(i)} | z_{1:T}^{(i)}; \Theta_{0:T}^{(i)}) \left[ \prod_{t=T+1}^{T+\tau} p(z_t^{(i)} | \mathbf{l}_{t}^{(i)}; \Theta_t^{(i)}) p(\mathbf{l}_t^{(i)} | \mathbf{l}_{t-1}^{(i)}; \Theta_t^{(i)}) \right] d\mathbf{l}_{T:T+\tau}^{(i)} .
\end{align}
Although the joint distribution is analytically tractable, it is evaluated by Monte Carlo approaches for the sake of implementation convenience following the reference paper. 

\section{Additional Results} \label{sec:ablation}
The mean and standard deviation value reported in following tables and figures are calculated over hyperparameter samples.

\subsection{DeepAR} \label{sec:deepar_hyper}
To illsutrate the effect of important hyperparameters to the performance of DeepAR, we report the result with different scaling methods, output distribution, context length, and number of RNN layers.
As reported in Section \ref{sec:deepar_results}, applying scaling method is essential to achieve desirable performance.
The longer the context length is, the better the RMSE and the mean NQL are.
The output distribution and number of RNN layers slightly varies the performance: negative binomial distribution is shown to be better fits to our EC dataset, and the performance drops as the number of layers increases probably due to the overfitting.

\subsection{DeepState} \label{sec:deepstate_hyper}
Primarily, we address the performance of DeepState with fully-learnable (FL) SSMs and partially-learnable (PL) SSMs.
We perform experiments with different context lengths and report each metrics's best mean score in Table \ref{tab:flpl} and Fig. \ref{fig:flpl}.
The performance of both models approximately converges after 56 days of context lengths.
The model with FL SSMs shows inaccurate and imprecise performance, and this appears to be a problem of the inherent lack of representation power in the model and non-identifiability problem \cite{doerr2018probabilistic, frigola2014variational} caused by SSMs' extreme flexibility.

Secondly, we compare the DeepState with different scaling methods: mean and median scalers.
The performance degrades significantly when the scaler is not used.
Empirically found that the model without scaler occasionally raise error when it calls Cholesky decomposition operator in Kalman filtering.

% \newpage
\vspace{1cm}

\begin{table}[h]
  \label{tab:scale}
  \caption{RMSE, MAPE, and Mean NQL by scaling method for DeepAR and DeepState models.}
  \centering
  \begin{tabular}{llll}
    \toprule
    Model       & RMSE      & MAPE    & Mean NQL[0.1, 0.9] \\
    \midrule
    DeepAR w/ mean scaler      & 81.37$\pm$2.60  & 0.303$\pm$0.013  & 0.140$\pm$0.008   \\
    DeepAR w/ median scaler  & \bf 81.06$\pm$1.07  & \bf 0.300$\pm$0.008  & \bf 0.139$\pm$0.006   \\
    DeepAR w/o scaler  & 94.08$\pm$1.59  & 0.308$\pm$0.005  & 0.162$\pm$0.007   \\
    \midrule
    DeepState w/ mean Scaler     & \bf 87.88 $\pm$ 11.00 & \bf 0.342 $\pm$ 0.022 & \bf 0.169 $\pm$ 0.028 \\
    DeepState w/ median scaler     & 114.68 $\pm$ 124.45 & 0.365 $\pm$ 0.118 & 0.211 $\pm$ 0.140 \\
    DeepState w/o scaler     & 150.91 $\pm$ 101.84 & 0.671 $\pm$ 0.297 & 0.427 $\pm$ 0.238 \\
  \bottomrule
  \end{tabular}
\end{table}

\begin{table}[h]
  \label{tab:layers}
  \caption{RMSE, MAPE, and Mean NQL by scaling method for DeepAR and DeepState models.}
  \centering
  \begin{tabular}{llll}
    \toprule
    Model       & RMSE      & MAPE    & Mean NQL[0.1, 0.9] \\
    \midrule
    DeepAR w/ 3-layer GRU     & 80.21 $\pm$ 0.34 & \bf 0.288 $\pm$ 0.002 & \bf 0.135 $\pm$ 0.002 \\
    DeepAR w/ 5-layer GRU     & \bf 79.91 $\pm$ 0.40 & 0.294 $\pm$ 0.007 & 0.136 $\pm$ 0.001 \\
    \midrule
    DeepState w/ 3-layer GRU     & 120.45 $\pm$ 89.05 & 0.467 $\pm$ 0.268 & 0.271 $\pm$ 0.210 \\
    DeepState w/ 5-layer GRU     & \bf 115.20 $\pm$ 103.49 & \bf 0.450 $\pm$ 0.203 & \bf 0.266 $\pm$ 0.182 \\
    \bottomrule
  \end{tabular}
\end{table}

\begin{table}[h]
  \label{tab:outprob}
  \caption{RMSE, MAPE, and Mean NQL by output distribution for DeepAR models.}
  \centering
  \begin{tabular}{llll}
    \toprule
    Model       & RMSE      & MAPE    & Mean NQL[0.1, 0.9] \\
    \midrule
    DeepAR w/ Gaussian output      & \bf 80.50$\pm$0.87  & 0.297$\pm$0.004  & 0.140$\pm$0.006   \\
    DeepAR w/ Negative Binomial output  & 80.87$\pm$1.27  & \bf 0.296$\pm$0.008  & \bf 0.138$\pm$0.006   \\
    \bottomrule
  \end{tabular}
\end{table}

\begin{table}[h]
  \label{tab:flpl}
  \caption{RMSE, MAPE, and Mean NQL by SSM architecture for DeepState models.}
  \centering
  \begin{tabular}{llll}
    \toprule
    Model       & RMSE      & MAPE    & Mean NQL[0.1, 0.9] \\
    \midrule
    DeepState w/ partially-learnable SSMs     & \bf 93.83 $\pm$ 11.51 & \bf 0.373 $\pm$ 0.077 & \bf 0.200 $\pm$ 0.042 \\
    DeepState w/ fully-learnable SSMs     & 141.82 $\pm$ 131.80 & 0.545 $\pm$ 0.304 & 0.338 $\pm$ 0.256 \\
    \bottomrule
  \end{tabular}
\end{table}

\begin{table}[h]
  \caption{RMSE, MAPE, and Mean NQL by the seasonality modes for Prophet models.}
  \centering
  \begin{tabular}{llll}
    \toprule
    Model       & RMSE      & MAPE    & Mean NQL[0.1, 0.9] \\
    \midrule
    Prophet w/ additive seasonality      & \bf 83.89$\pm$1.98  &  0.461$\pm$0.038  & \bf 0.202$\pm$0.009   \\
    Prophet w/ multiplicative seasonality & 97.02$\pm$16.85  & \bf 0.460$\pm$0.044  & 0.222$\pm$0.010   \\
    \bottomrule
  \end{tabular}
\end{table}

\begin{figure}[h]
    \centering
    \subfigure[RMSE]{
        \includegraphics[width=0.31\textwidth]{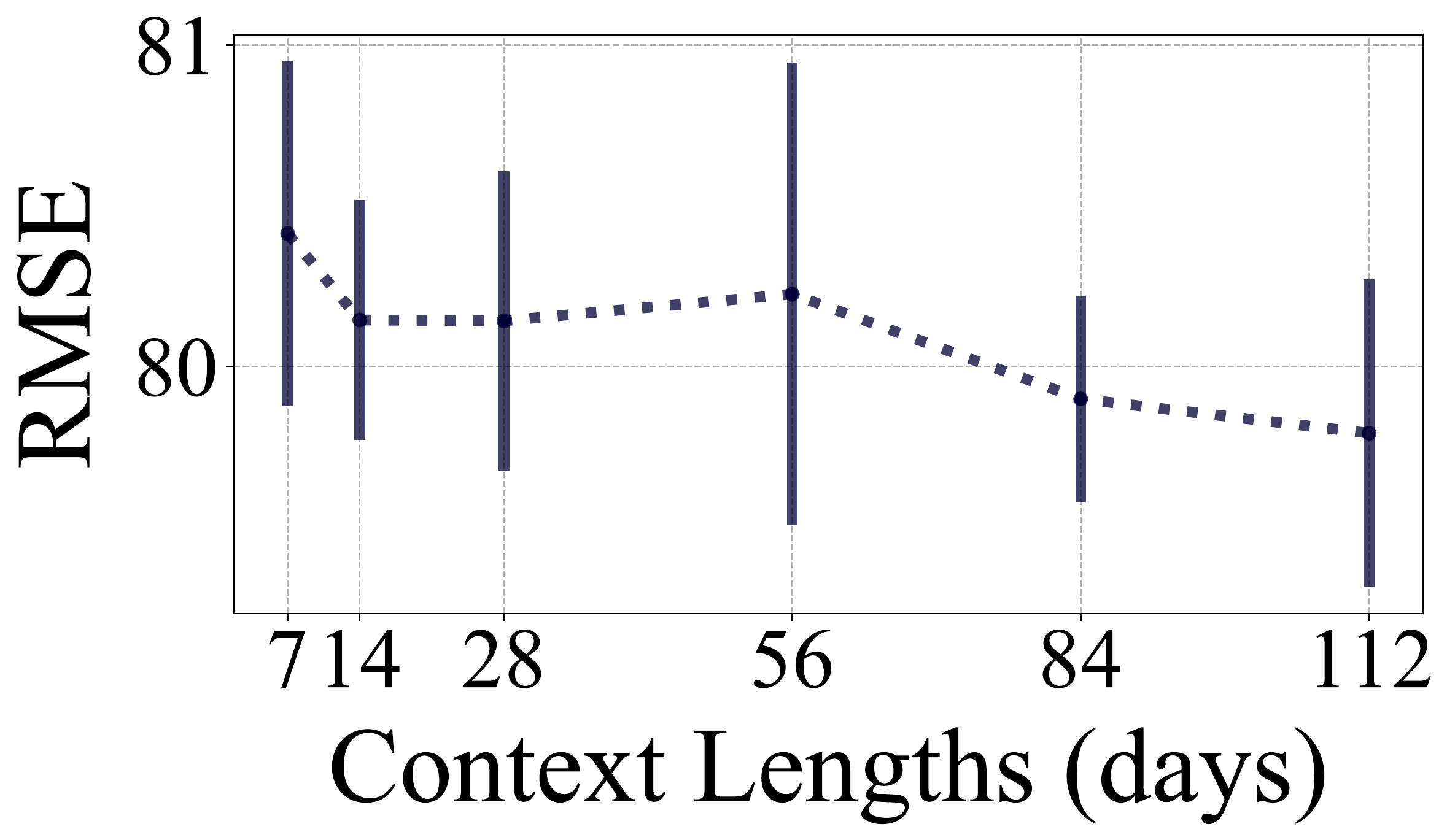}
    }
    \subfigure[MAPE]{
        \includegraphics[width=0.31\textwidth]{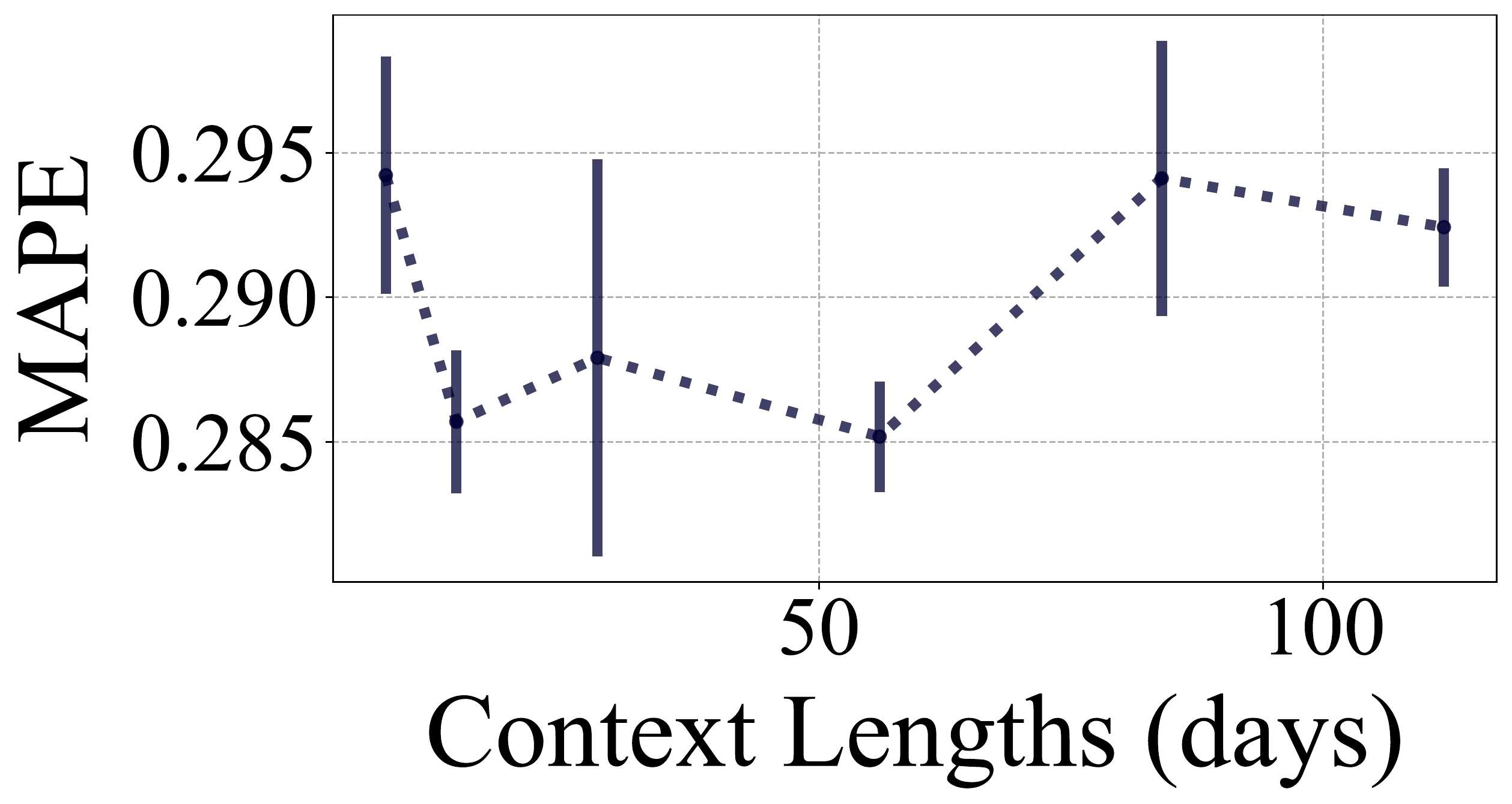}
    }
    \subfigure[{Mean NQL[0.1, 0.9]}]{
        \includegraphics[width=0.31\textwidth]{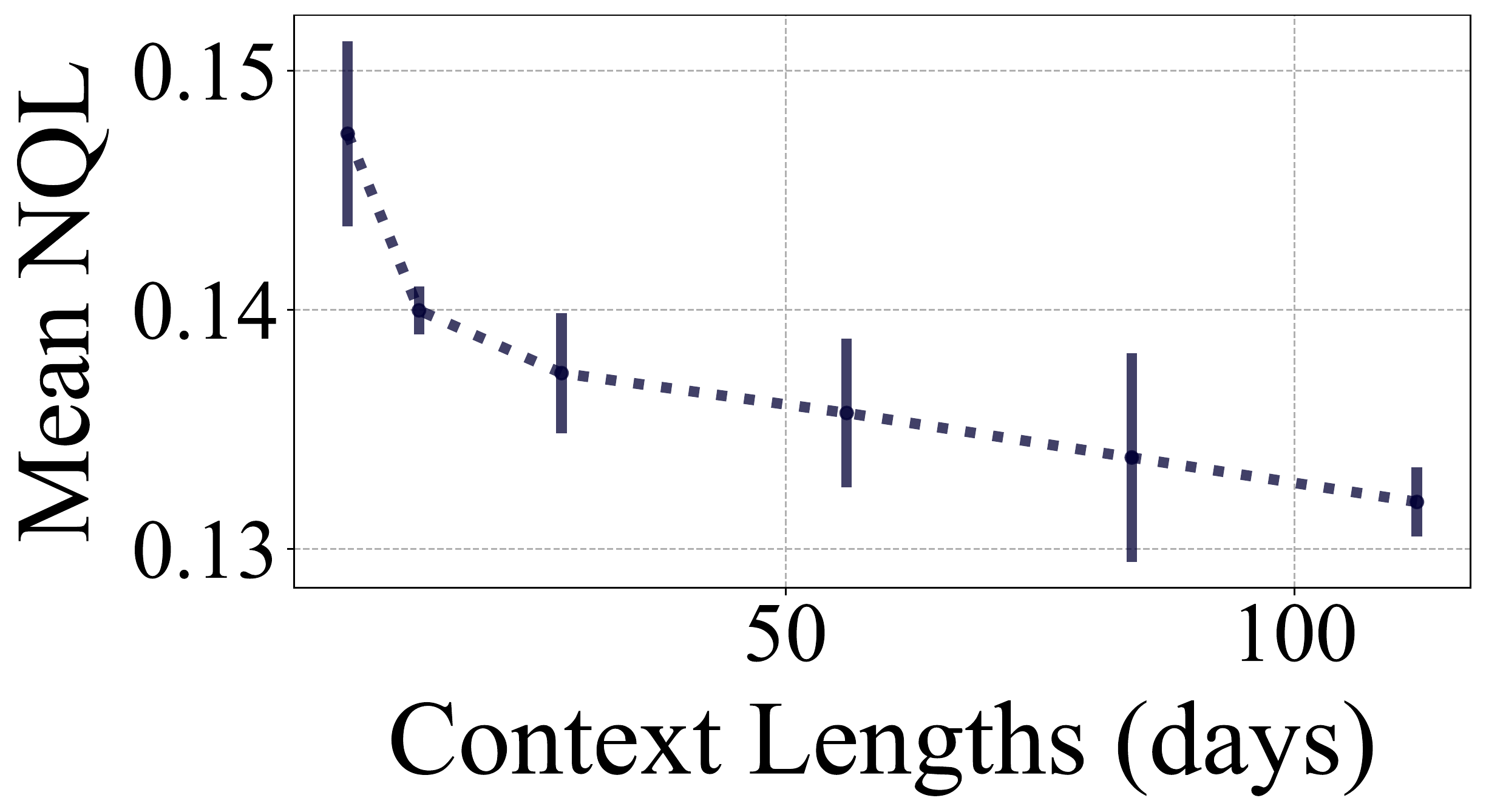}
    }
    \caption{RMSE, MAPE, and Mean NQL by the context lengths for DeepAR models.}
\end{figure}

\begin{figure}[h]
    \label{fig:flpl}
    \centering
    \subfigure[RMSE]{
        \includegraphics[width=0.31\textwidth]{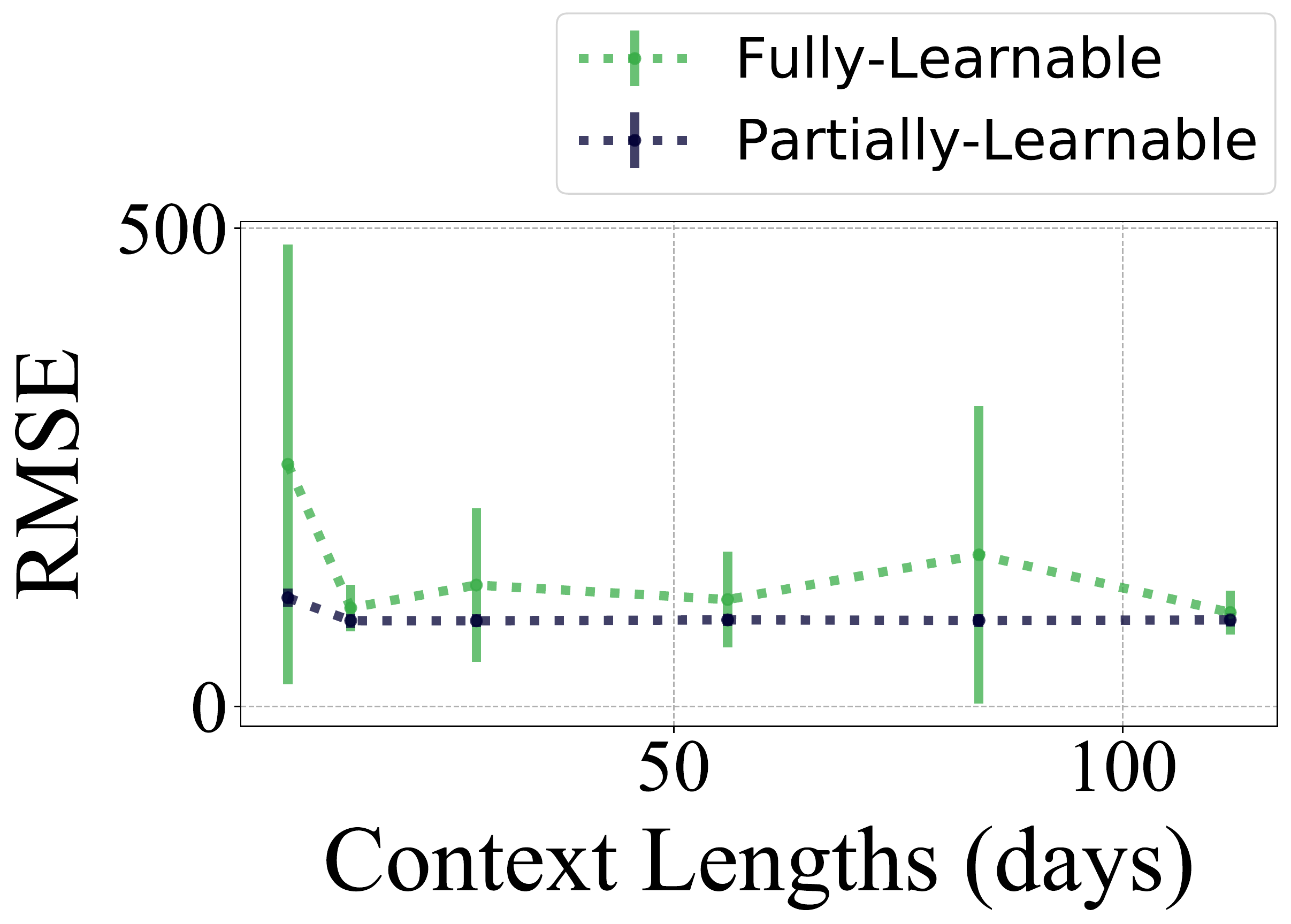}
    }
    \subfigure[MAPE]{
        \includegraphics[width=0.31\textwidth]{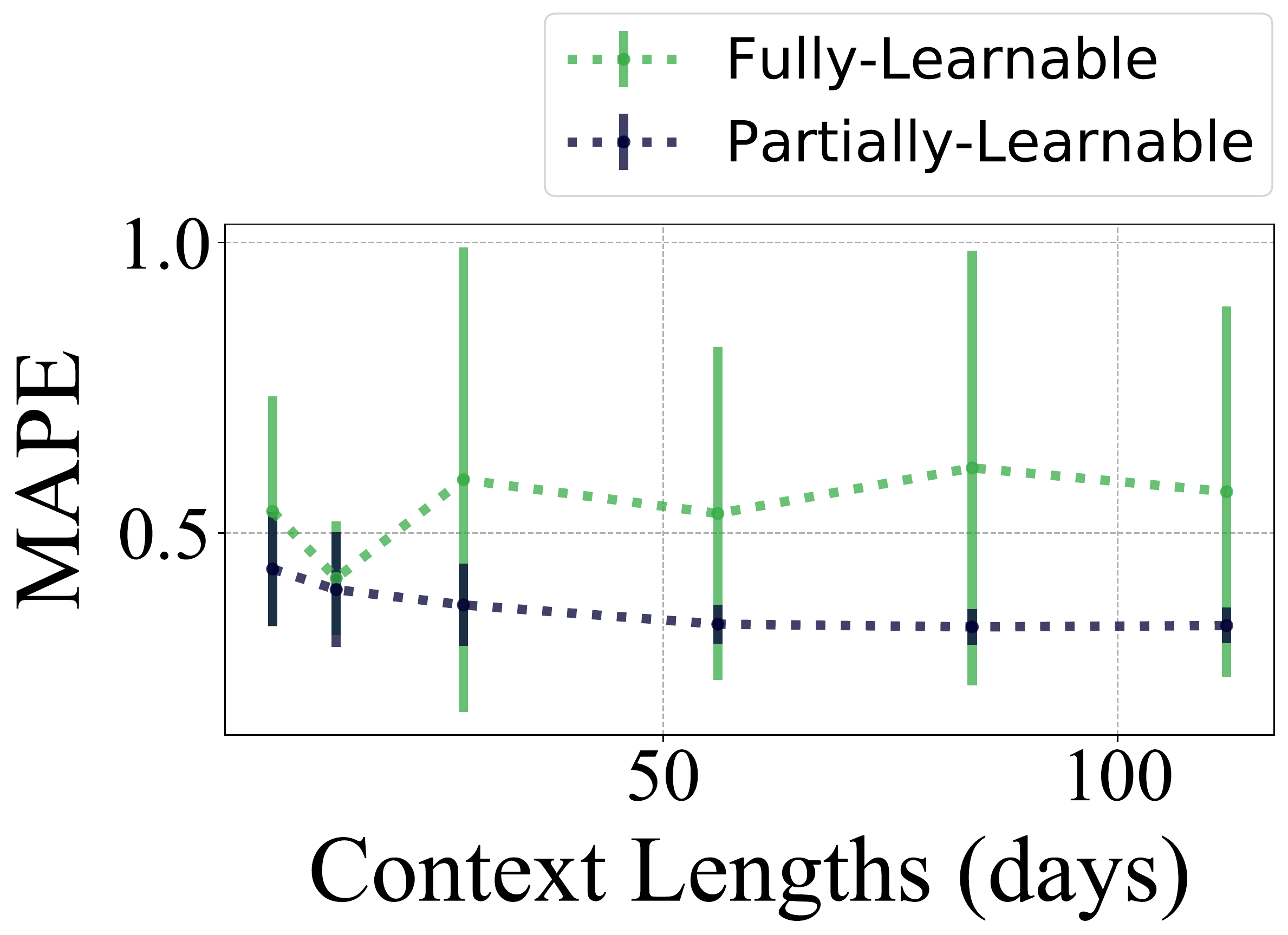}
    }
    \subfigure[{Mean NQL[0.1, 0.9]}]{
        \includegraphics[width=0.31\textwidth]{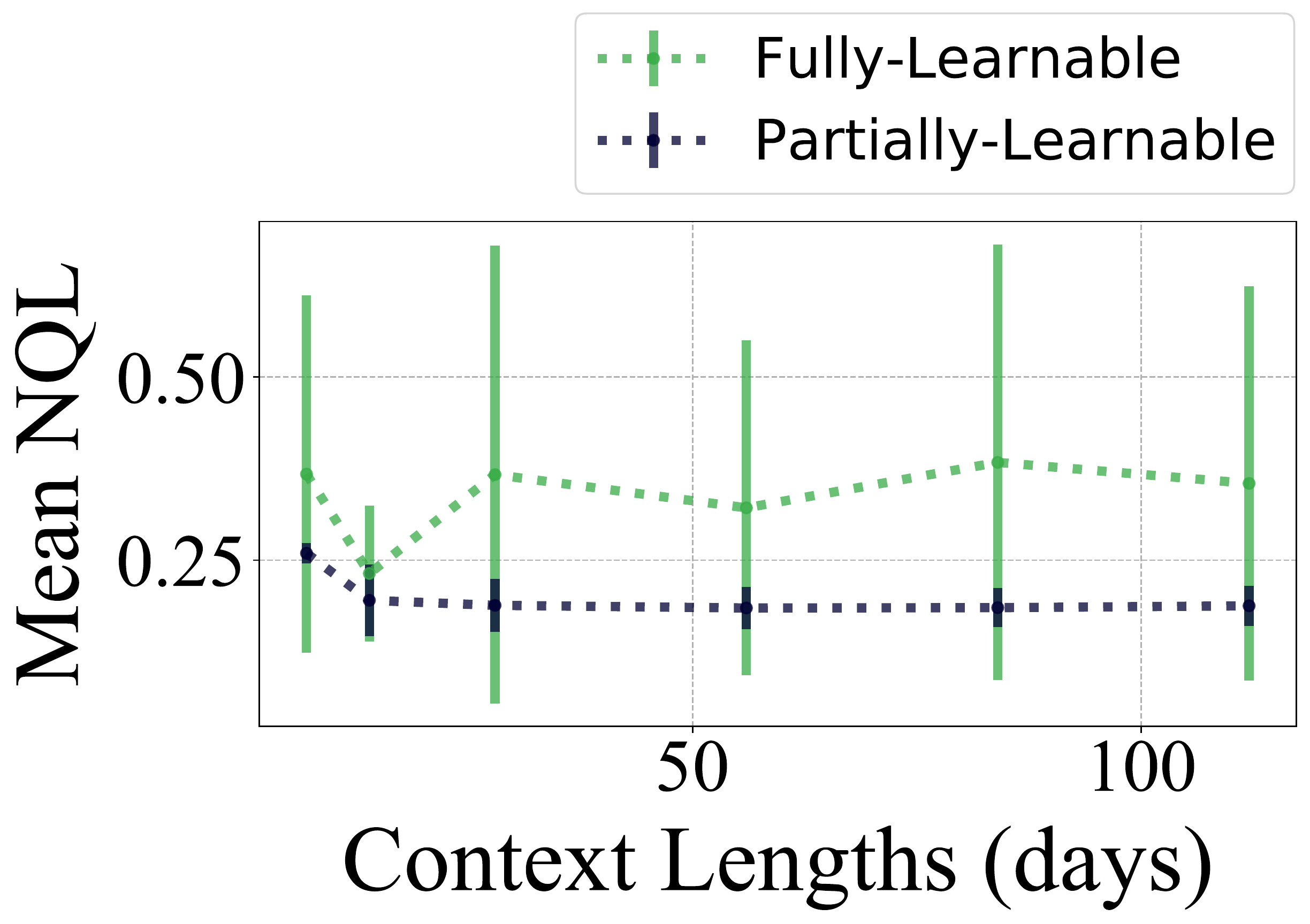}
    }
    \caption{RMSE, MAPE, and Mean NQL by the context lengths for DeepState models.}
\end{figure}

\begin{figure}[h]
    \centering
    \subfigure[RMSE]{
        \includegraphics[width=0.31\textwidth]{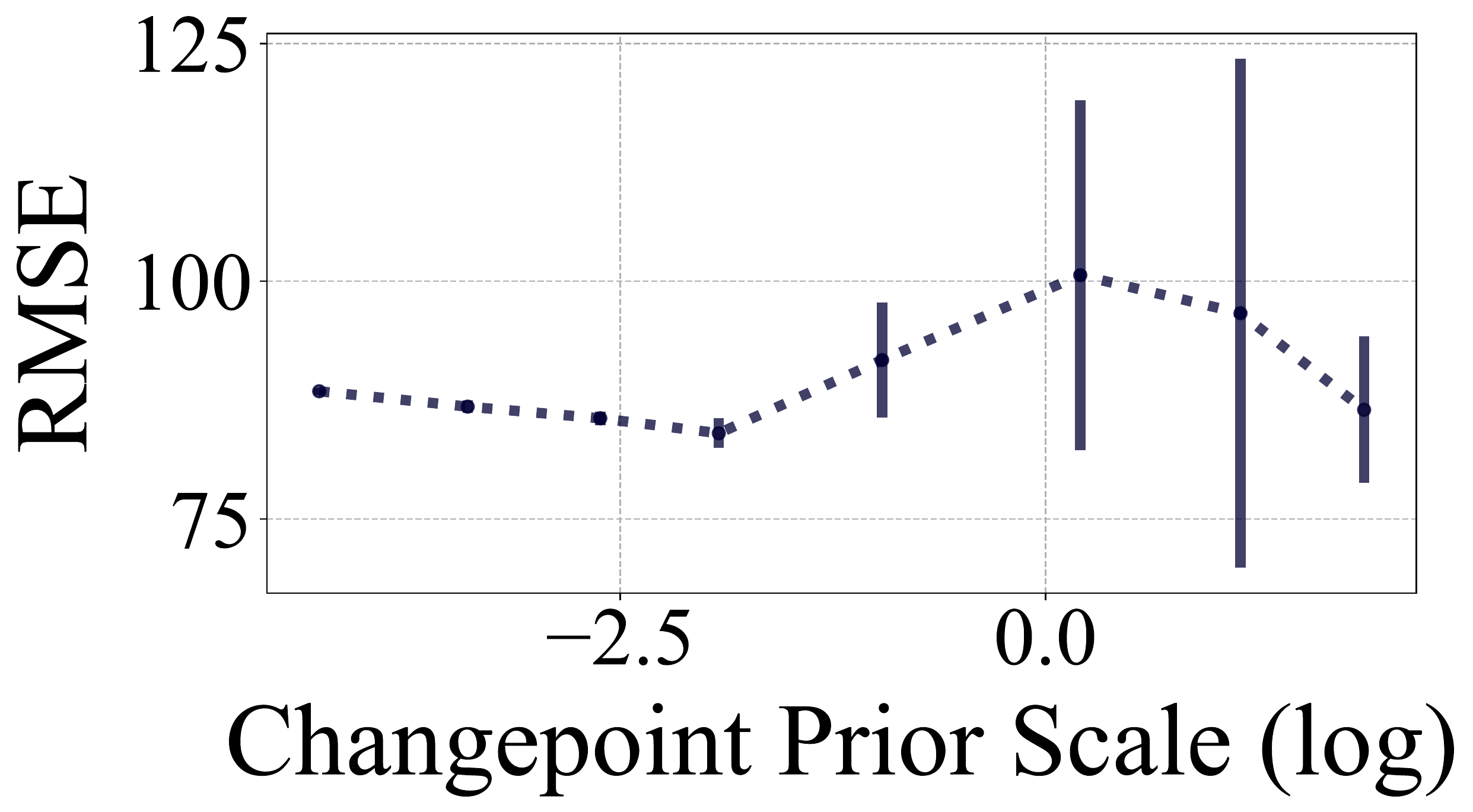}
    }
    \subfigure[MAPE]{
        \includegraphics[width=0.31\textwidth]{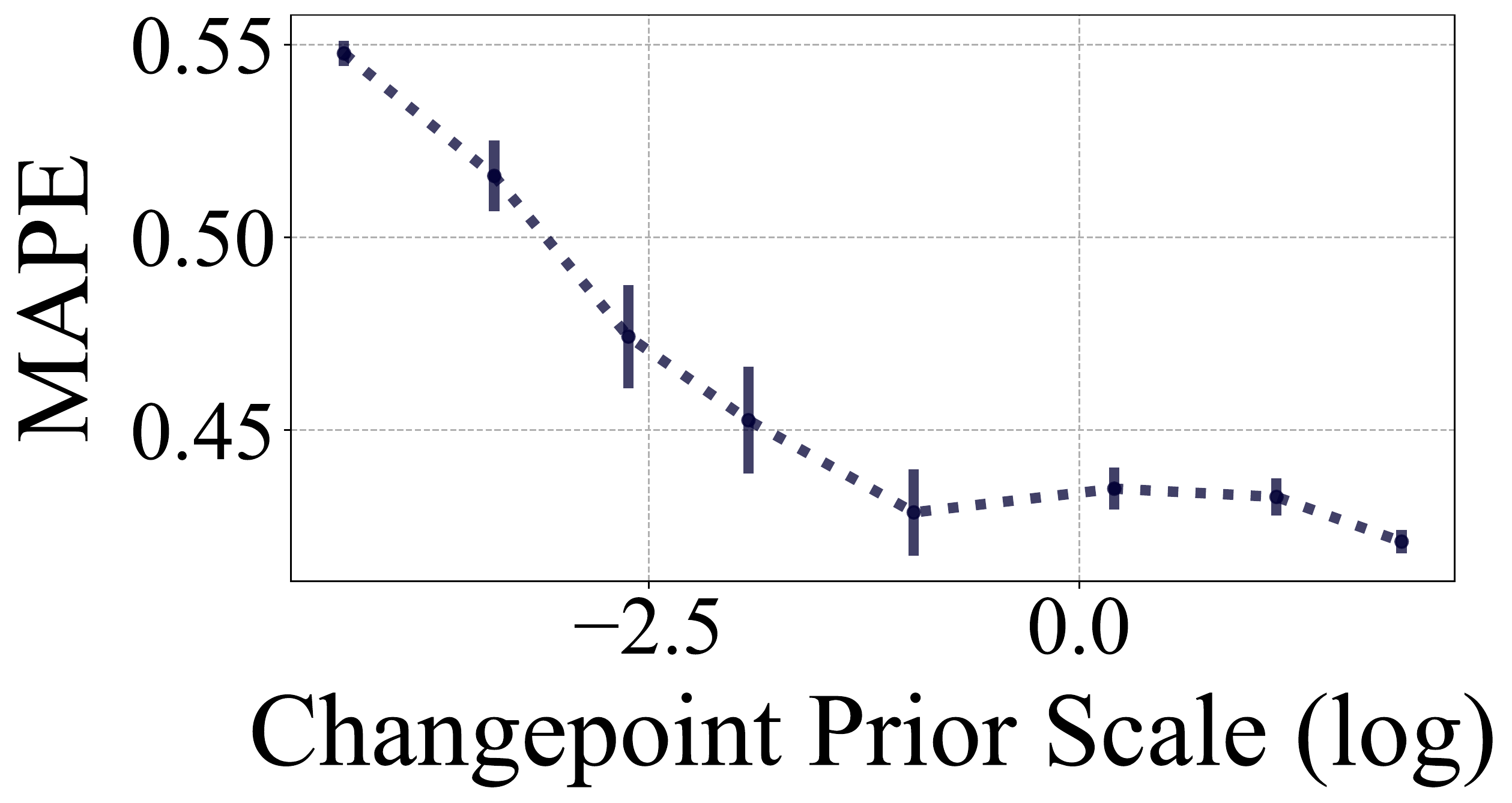}
    }
    \subfigure[{Mean NQL[0.1, 0.9]}]{
        \includegraphics[width=0.31\textwidth]{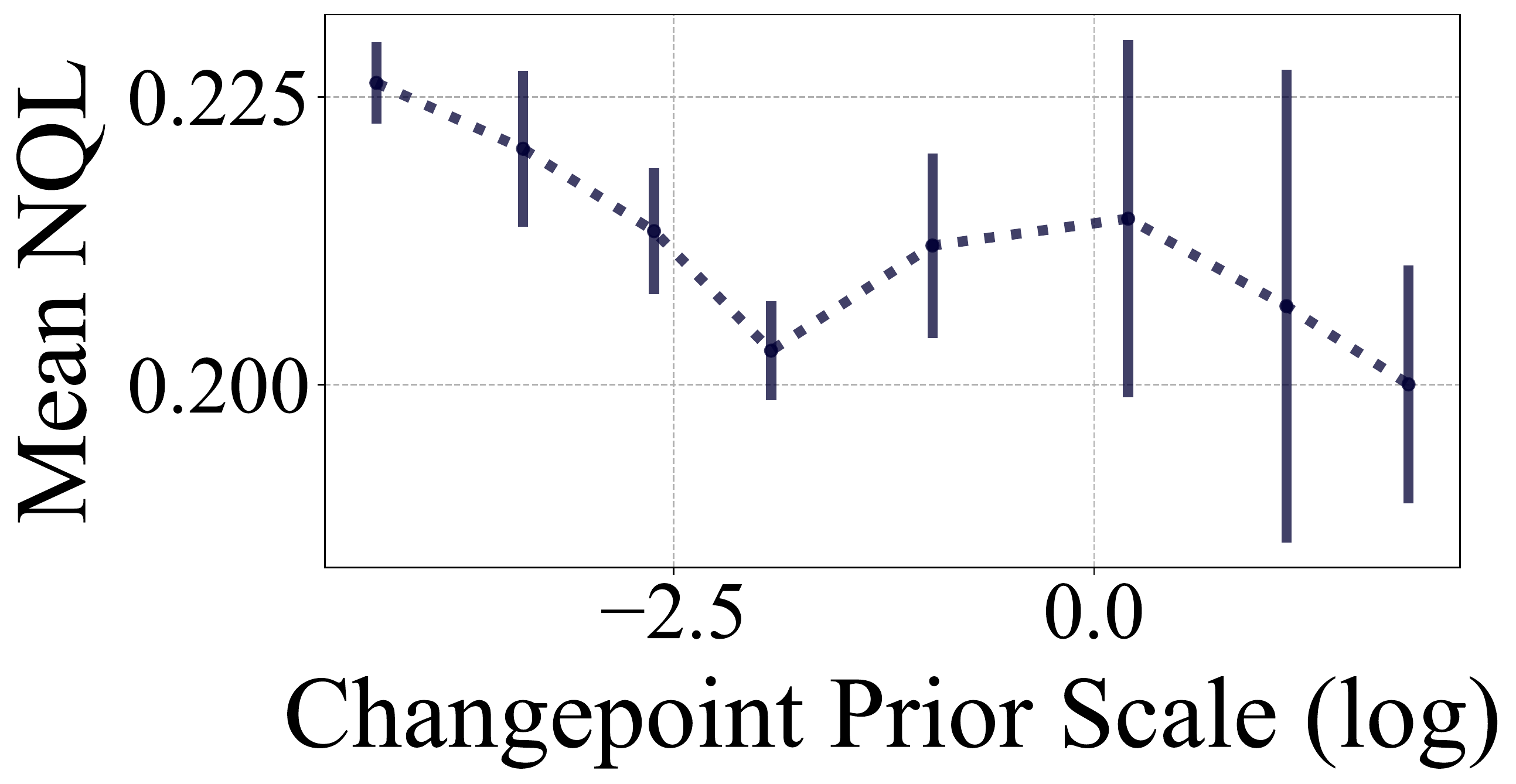}
    }
    \caption{RMSE, MAPE, and Mean NQL by the changepoint prior scales for Prophet models.}
\end{figure}

\begin{figure}[h]
    \centering
    \subfigure[RMSE]{
        \includegraphics[width=0.31\textwidth]{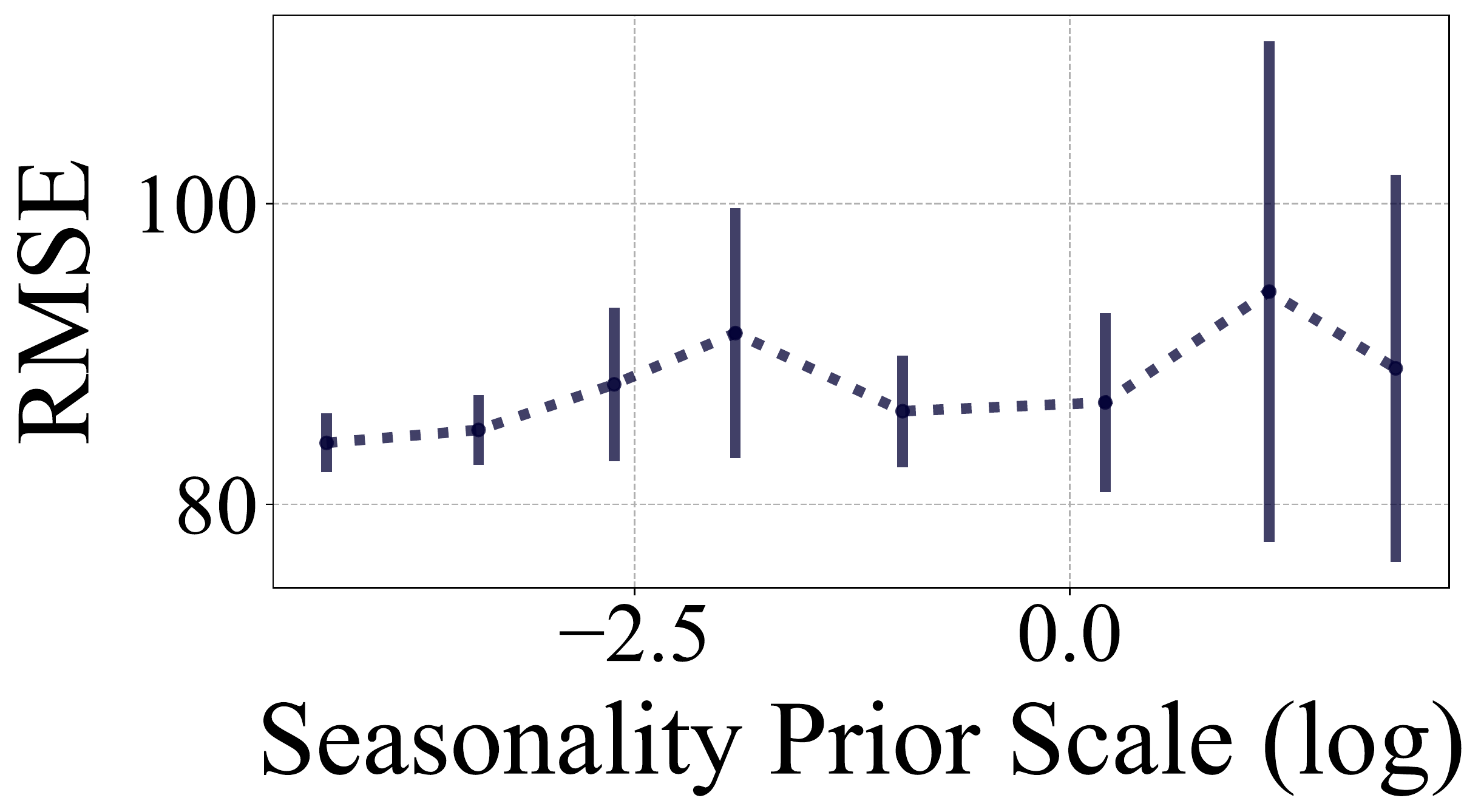}
    }
    \subfigure[MAPE]{
        \includegraphics[width=0.31\textwidth]{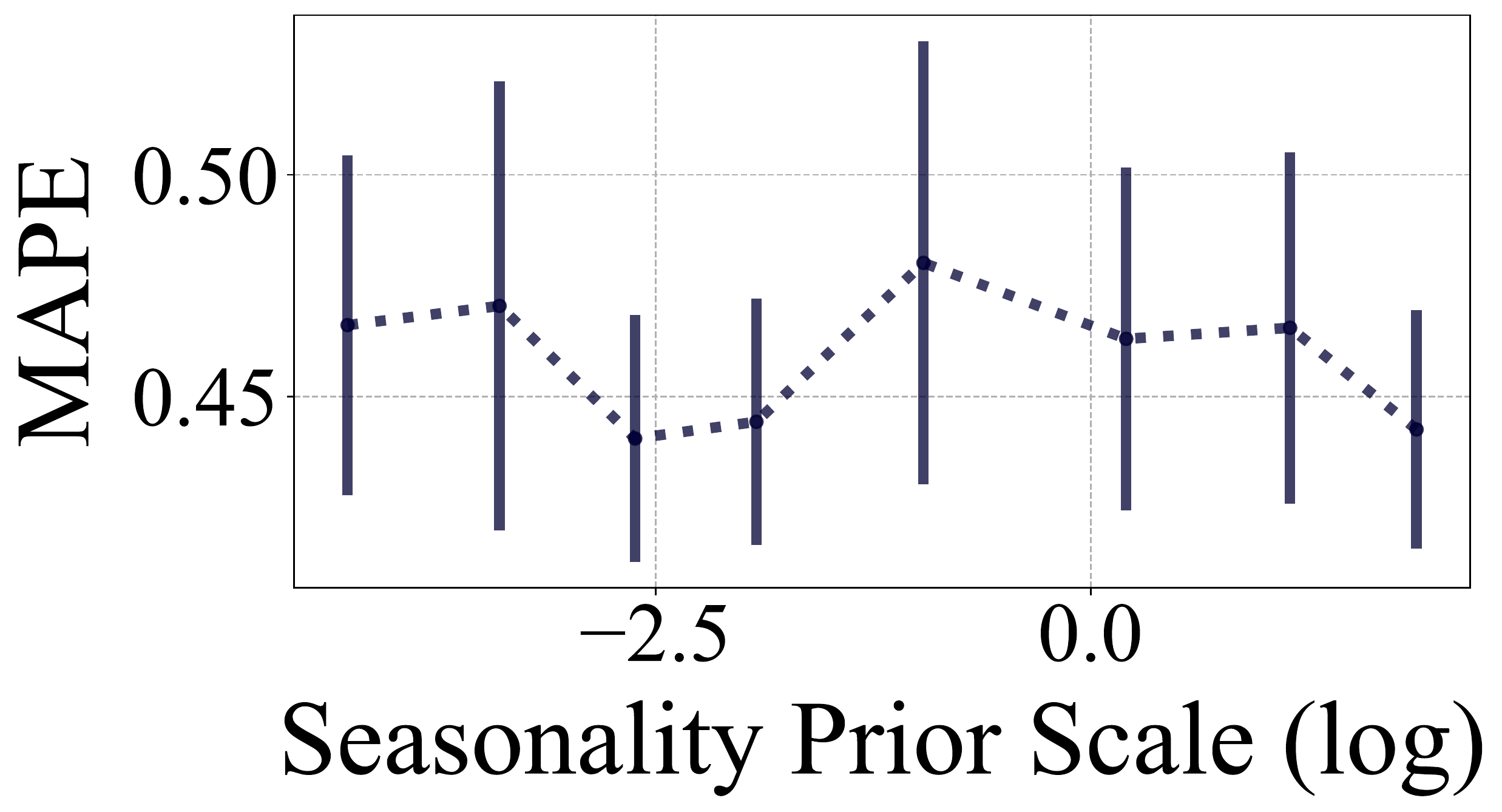}
    }
    \subfigure[{Mean NQL[0.1, 0.9]}]{
        \includegraphics[width=0.31\textwidth]{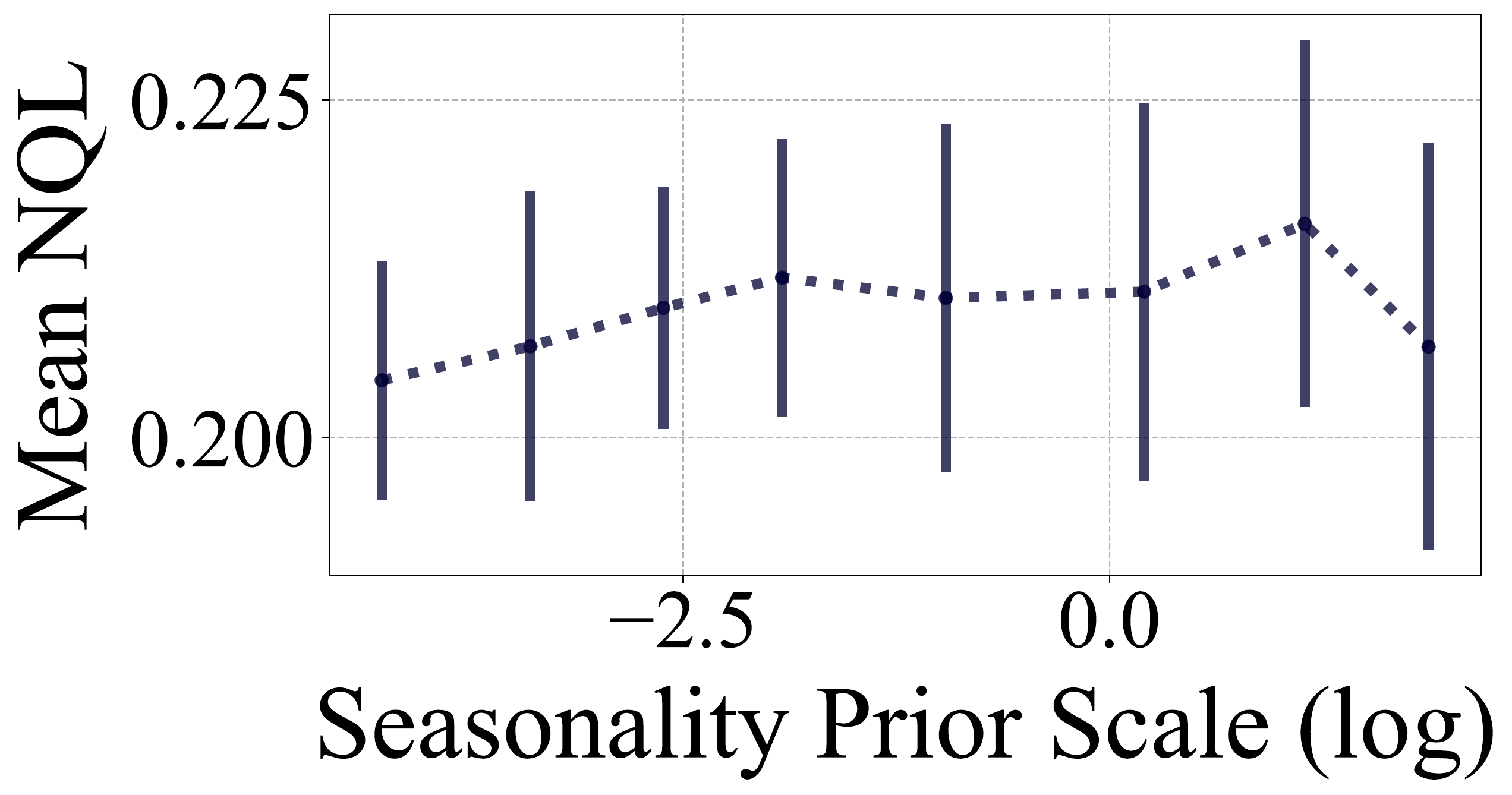}
    }
    \caption{RMSE, MAPE, and Mean NQL by the seasonality prior scales for Prophet models.}
\end{figure}

\end{document}